\documentclass{article}

\usepackage[preprint]{neurips_2024}

\usepackage[utf8]{inputenc} 
\usepackage[T1]{fontenc}    
\usepackage{hyperref}       
\usepackage{url}            
\usepackage{booktabs}       
\usepackage{amsfonts}       
\usepackage{nicefra  c}       
\usepackage{microtype}      
\usepackage{xcolor}         
\usepackage{tcolorbox}
\usepackage{bbding}
\usepackage{graphicx}
\usepackage{amsmath}
\usepackage{amssymb}
\usepackage{multirow}
\usepackage{color}
\usepackage{threeparttable}
\usepackage{colortbl} 
\usepackage{makecell}
\usepackage{tabularx}
\usepackage{caption}
\usepackage{subcaption}
\usepackage{colortbl} 
\usepackage{ulem}
\usepackage{authblk}

\title{Multi-Granularity Semantic Revision for Large Language Model Distillation
}

\author[1,*]{Xiaoyu Liu}
\author[2,\thanks{Equal Contribution.\quad$^\ddagger$ Project Leader. \quad$\dagger$Corresponding Author.}]{Yun Zhang}
\author[3,$^\ddagger$]{Wei Li}
\author[3]{Simiao Li}
\author[3]{Xudong Huang}
\author[3]{Hanting Chen}
\author[3]{Yehui Tang}
\author[3]{Jie Hu}
\author[1]{Zhiwei Xiong}
\author[3,$^\dagger$]{Yunhe Wang}
\affil[1]{University of Science and Technology of China}
\affil[2]{DSA Thrust, INFO
Hub, Hong Kong University of Science and Technology (GZ)}
\affil[3]{Huawei Noah’s Ark Lab}
\affil[ ]{\tt\small liuxyu@mail.ustc.edu.cn, \{wei.lee, yunhe.wang\}@huawei.com}

\begin{document}

\maketitle

\begin{abstract}
Knowledge distillation plays a key role in compressing the Large Language Models (LLMs), which boosts a small-size student model under large teacher models' guidance. However, existing LLM distillation methods overly rely on student-generated outputs, which may introduce generation errors and misguide the distillation process. 
Moreover, the distillation loss functions introduced in previous art struggle to align the most informative part due to the complex distribution of LLMs' outputs.
To address these problems, we propose a multi-granularity semantic revision method for LLM distillation.  At the sequence level, we propose a sequence correction and re-generation (SCRG) strategy. 
SCRG first calculates the semantic cognitive difference between the teacher and student to detect the error token, then corrects it with the teacher-generated one, and re-generates the sequence to reduce generation errors and enhance generation diversity.
At the token level, we design a distribution adaptive clipping Kullback-Leibler (DAC-KL) loss as the distillation objective function. DAC-KL loss exploits a learnable sub-network to adaptively extract semantically dense areas from the teacher's output, avoiding the interference of redundant information in the distillation process. Finally, at the span level, we leverage the span priors of a sequence to compute the probability correlations within spans, and constrain the teacher and student's probability correlations to be consistent, further enhancing the transfer of semantic information. Extensive experiments across different model families with parameters ranging from 0.1B to 13B demonstrate the superiority of our method compared to existing methods.

\end{abstract}

\section{Introduction}

The remarkable advancements in auto-regressive Large Language Models (LLMs)~\cite{kaplan2020scaling,Wei_2022,Radford_Wu_Child_Luan_Amodei_Sutskever,Zhangetal,Brown_Mann} have led to unprecedented breakthroughs in a diverse array of text generative tasks, with numerous open-source models~\cite{Touvron,zhang2022opt} now available. A crucial factor contributing to this success is the ability to scale up the models, which involves increasing both the amount of training data and the number of model parameters. However, the massive size and computational intensity of these state-of-the-art models pose significant challenges, particularly when it comes to deployment and real-time applications. In contrast, smaller models with limited parameters often sacrifice performance on real-world generation tasks~\cite{wang2022self}. To mitigate these challenges, Knowledge Distillation (KD)~\cite{hinton2015distilling} has emerged as a pivotal technique, enabling the development of smaller, more efficient student models that inherit the strengths of their larger teacher counterparts.

\begin{figure}[!ht]
  \centering
   \includegraphics[width=\textwidth]{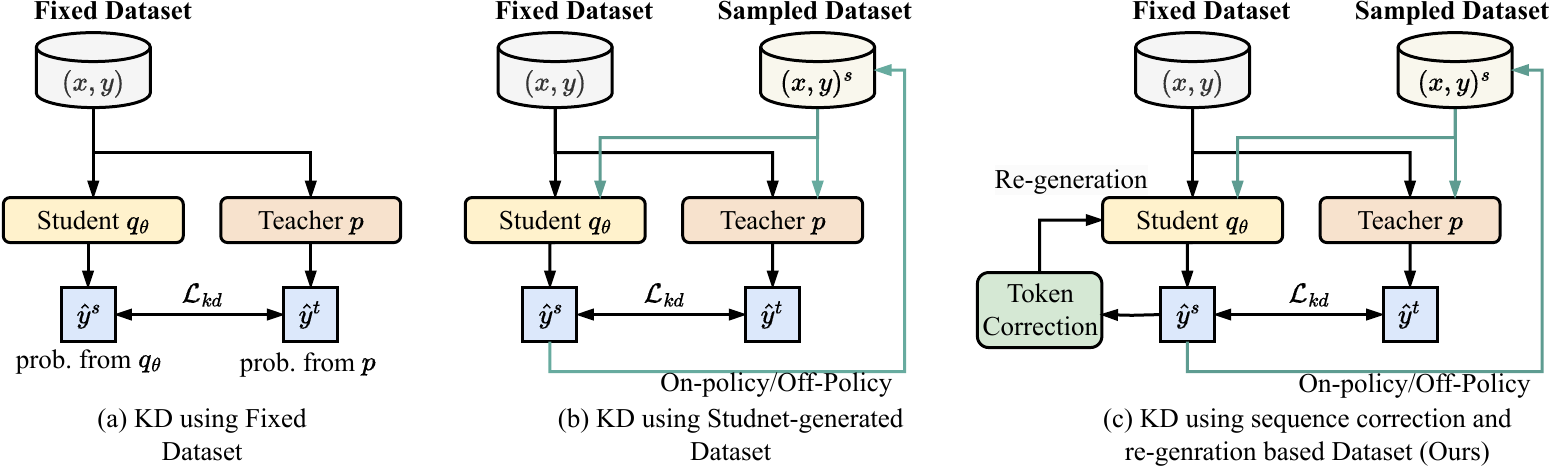} %
\caption{Knowledge Distillation using Different Sampled Datasets. (a) Traditional KD using a fixed dataset~\cite{hinton2015distilling}. (b) KD using the student-generated dataset, which can be categorized into on-policy based methods~\cite{agarwal2024policy,gu2023minillm} and the off-policy based method~\cite{ko2024distillm}. (c) Our proposed KD approach, which leverages a sequence correction and re-generation strategy and can be seamlessly integrated with both on-policy and off-policy generation schedules.}
  \label{generation_comparison}
\vspace{-0.5cm}
\end{figure}

Traditional knowledge distillation methods~\cite{hinton2015distilling, kim2016sequence} directly employ Kullback-Leibler divergence (KLD) as the distillation loss for aligning the output distributions of teacher and student models on a static dataset (see Figure~\ref{generation_comparison} (a)). Unlike these methods, recent LLM distillation methods are exploring diverse divergence loss functions tailored to LLMs and leveraging student-generated datasets to avoid distribution mismatch between the outputs student-generated in the training and inference stages. GKD~\cite{agarwal2024policy} and MiniLLM~\cite{gu2023minillm} propose to exploit reverse KLD as the distillation objective, replacing the commonly used forward KLD. These approaches aim to prevent students from overestimating the low-probability regions of the teacher's distribution. Also, these methods train the student on self-generated sequences that are on-policy instead of a fixed set of output sequences. Recently, Distillm~\cite{ko2024distillm} proposes an adaptive off-policy student-generation strategy to improve the sample efficiency and high generation time faced in on-policy generation (see Figure~\ref{generation_comparison} (b)). Meanwhile, it designs a new distillation object function \textit{i.e.}, skew KLD loss for better generalizability and convergence. 
However, relying on student-generated sequences may introduce generation errors and lead to suboptimal learning, as the distillation process becomes vulnerable to the inaccuracies inherent in the student's predictions. The student model's limited capacity and biases can further perpetuate these errors, resulting in a distorted representation of the teacher's knowledge. Moreover, the rich semantic knowledge and the significant variance across different tokens make it challenging for existing distillation objective functions to capture and transfer the essential knowledge within the teacher model's output distribution.

To address the above-mentioned issues, we introduce a novel multi-level semantic revision approach, across sequence token and span levels, to significantly improve the KD process for LLMs. 
At the sequence level, we propose a sequence correction and re-generation (SCRG) strategy. We detect the error token in the student-generated sequence and re-generate the sequence from the position of the error token to reduce generation errors and enhance generation diversity. As shown in Figure~\ref{generation_comparison} (c), by assessing the semantic cognitive differences between teacher and student outputs on a token-by-token basis, we identify and correct errors, leading to re-generated sequences that steer the student model towards generating more reliable and diverse samples and can be seamlessly integrated with both on-policy and off-policy generation schedules.
At the token level, we employ a distribution adaptive clipping Kullback-Leibler (DAC-KL) loss function, which leverages a learnable sub-network to target semantically salient regions of the output distribution. This loss function effectively filters out redundant information, preserving only the most relevant signals for distillation.
Finally, at the span level, we incorporate pre-defined span priors of sequences to align the relations of probability vectors of the student and teacher models, ensuring a consistent transfer of semantic information across related tokens within the same span.
Through extensive experiments with different models, including the LLAMA2, OpenLLAMA2, OPT, and GPT2 series, ranging from 0.1B to 13B parameters, we showcase the superiority of our approach over existing knowledge distillation methods.

The contributions of this paper are summarized as follows:

\begin{itemize} \itemsep -2pt
\item We introduce a novel multi-level semantic revision approach to enhance the knowledge distillation (KD) process for large language models (LLMs).
\item At the sequence level, we propose a sequence correction and re-generation strategy to steer the student model towards generating more reliable and diverse sequences.
\item At the token level, we propose a distribution adaptive clipping Kullback-Leibler loss to capture semantically salient regions of the output space.
\item At the span level, we incorporate input span priors to ensure a consistent transfer of semantic knowledge across related tokens.
\item Through extensive experimentation with models ranging from 0.1B to 13B parameters, we demonstrate the superiority of our method over existing KD methods for LLMs.
\end{itemize}

\section{Related work}
\paragraph{KD for encoder-only language models.}
Pretrained encoder-only language models, such as BERT~\cite{jiao2019tinybert}, can be compressed using the traditional logit distillation~\cite{hinton2015distilling} and feature distillation~\cite{adriana2015fitnets}. These knowledge distillation methods minimize the Kullback-Leibler divergence loss between the outputs of the student and teacher models on a fixed dataset~\cite{kim2016sequence}. Liang et al.\cite{liang2020mixkd} applied this objective to train students on masked language modelling and text classification tasks. Jiao et al.\cite{jiao2019tinybert} utilized intermediate representations from each transformer layer of the teacher as transferable knowledge. Despite the potential of KD in encoder-only language models~\cite{sanh2019distilbert,liang2023less,sun2019patient,liu2022multi}, the complex predictions generated by large language models (LLMs) through auto-regressive inference present a new challenge. This paper primarily discusses KD for auto-regressive LLMs.

\paragraph{KD for auto-regression large language models.}

Existing knowledge distillation (KD) methods for auto-regressive large language models (LLMs) can be divided into black-box methods for closed-source models such as GPT-3.5~\cite{ouyang2022training} and GPT-4~\cite{achiam2023gpt}, and white-box methods for open-source models such as LLaMA~\cite{Touvron}. Black-box methods~\cite{chen2024knowledge,jiang2023lion,hsieh2023distilling} cannot access the internal parameters of the teacher model and utilize only the inference results provided by the teacher API~\cite{taori2023stanford,chiang2023vicuna,peng2023instruction}. The inference results of the teacher model are used to construct prompt-response pairs, which serve as a new training dataset to fine-tune the student model. In contrast, white-box KD methods~\cite{ko2024distillm,agarwal2024policy,gu2023minillm} leverage the internal parameters of the teacher model, providing richer training signals such as the probability distribution of predictions, potentially leading to better student model performance. Our methods primarily address the challenges of existing methods in the realm of white-box KD.

\section{Preliminary}

Before introducing our method, we provide some preliminary information on KD for LLMs. We consider the inference of LLMs as a vocabulary classification task, where a model $p$ predicts the conditional probability distribution of a target response $y$ given a prompt and target sequence pair $(x,y)$. Let $y_{<i}=(y_1, y_2,....,y_{i-1})$ denote the generated output sequence up to the ${(i-1)}^{th}$ token $y_{i-1}$. A token-level auto-regression model outputs a next-token $M-$vocabulary probability distribution. Specifically, for the model $p$,  $\hat{y}_i= p(.|y_{<i},X) (\hat{y}_i\in\mathbb{R}^M)$ represents the probability distribution of the generated $i^{th}$ token, where $\hat{y}_i \in (0,1)^M$. $y_i\sim p(.|y_{<i},X)$ is the corresponding output token.

We formulate KD as an optimization problem that aims to minimize the difference between the prediction distribution of a fixed teacher model $p(.|y_{<i},x)$ and that of a parameterized student model $q_\theta(.|y_{<i},x)$, using sampled input-output sequence pairs ($x$,$y$) from the fixed dataset ($X$,$Y$). $\theta$ is the student's parameters to be optimized. The sequence-level distillation with $L_y$ tokens employs KL Divergence $D_{KLD}$ as the distillation object.  The total distillation loss $\mathcal{L}_{KD}$ is broken down into a sum of token-wise distillation:
\begin{equation}
\label{kd_eq}
\begin{aligned}
\mathcal{L}_{KD} =\tfrac{1}{L_y}\sum_{i=1}^{L_y}D_{KLD}(p(.|y_{<i},x)||q_\theta(.|y_{<i},x))
=\tfrac{1}{L_y}\sum_{i=1}^{L_y}p(.|y_{<i},x)log\frac{p(.|y_{<i},x)}{q_\theta(.|y_{<i},x))},
\end{aligned}
\end{equation}
where the conditional sequence $y$ can be easily generated by sampling from the teacher or student model policy, \textit{i.e.},$\{x\in X, y\sim p(.|x)\}$ or $\{x\in X, y\sim q_\theta(.|x)\}$ instead of directly $\{(x,y)\in(X, Y)\}$.

During the distillation process, the student model is also guided by the ground-truth output sequence without querying the policies of the teacher or student models. The supervised fine-tuning (SFT) loss is formulated as
\begin{equation}
\begin{aligned}
\mathcal{L}_{\text{SFT}} =\mathbb{E}_{(x,y)\sim (X,Y)}[-log\ q_\theta(y|x) ].
\end{aligned}
\end{equation}

\section{Multi-Granularity Semantic Revision}

In this section, we introduce the proposed multi-granularity semantic revision for LLM distillation, which revises the semantic representation during the knowledge transfer stage at three levels: sequence-level, token-level, and span-level.

\subsection{Sequence-level correction and re-generation}
\begin{figure}[!t]
  \centering
   \includegraphics[width=\textwidth]{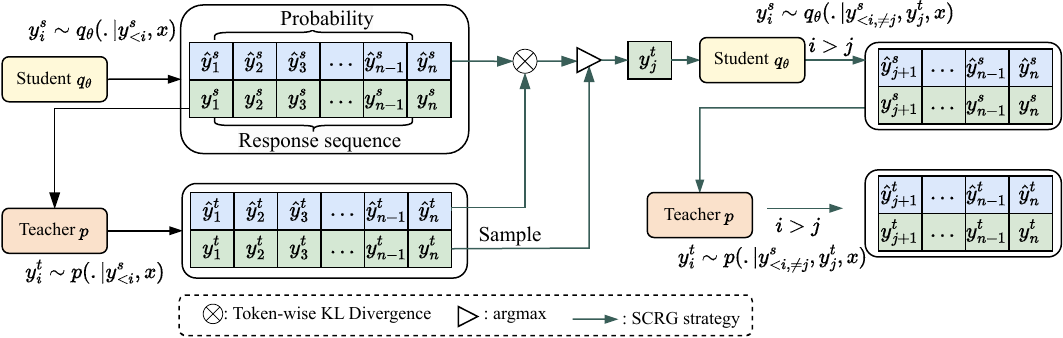} %
  \caption{The workflow of sequence correction and re-generation strategy.}
  \label{re-generation}
  \vspace{-0.4cm}
\end{figure}

As illustrated by Eq.~\eqref{kd_eq}, prevalent KD methods  \cite{agarwal2024policy,gu2023minillm,ko2024distillm}, utilizes conditional sequences generated from the student model (denoted as \( y \sim q_\theta(\cdot|x) \) ) for the distillation process. While these methods are designed to mitigate the training-inference mismatch between the fixed training data and the student's auto-regressive inferences, they simultaneously risk introducing generation errors. Due to the limited capabilities of the student model, the generated sequences may contain additional errors which reduces the effectiveness of KD. To address this issue, we propose a sequence correction and re-generation (SCRG) strategy (shown in Fig.~\ref{re-generation}) to detect generation errors and re-generate sequences that steer the student model towards generating reliable and diverse sequences. 

We denote the generated $n$-token sequence from the student model $q_\theta$ as $y^s_{<n+1} = (y^s_1, y^s_2,....,y^s_{n})$ which correspondences the probability outputs $(\hat{y}^s_1, \hat{y}^s_2,....,\hat{y}^s_{n})$, where $y^s_i \sim  q_\theta(.|y^s_{<i},x) (1\le i\le n)$. Similarly, we denote the teacher model's output sequence as $y^t_{<n+1} = (y^t_1, y^t_2,....,y^t_{n})$ and probability outputs $(\hat{y}^t_1, \hat{y}^t_2,....,\hat{y}^t_{n})$. We denote each token of the teacher model's output sequence as $y^t_i \sim p(.|y^s_{<i},x)$. We follow previous methods~\cite{agarwal2024policy,gu2023minillm,ko2024distillm} using the student-generated outputs as the distillation dataset, and calculate token-wise KLD loss to evaluate the semantic cognitive differences between the teacher and student for each token to detect the position of the error token within the sequence $y^t_{<n+1}$. We formulate the  detection process of the error token $y_j^s$ as
\begin{equation}
\begin{aligned}
j = \mathop{\arg\max}\limits_{1 \leq i \leq n} \left( \text{KLD}(\hat{y}^s_{i} \| \hat{y}^t_{i}) \ \text{if} \ y^s_{i} \ne y^t_{i} \right).
\end{aligned}
\end{equation}

We then replace the $y_j^s$ by $y_j^t$ to construct new samples and re-generate the student output sequence and each token in $y^s_{<n+1}$ is formulated as
\begin{equation}
\begin{aligned}
y^s_i \sim\begin{cases}
 q_\theta(.|y^s_{<i},x) & \text{ if } i<j \\
  p(.|y^s_{<i},x)& \text{ if } i=j \\
  q_\theta(.|y^s_{<i,\ne j},y^t_{j},x)& \text{ if } i>j.
\end{cases}
\end{aligned}
\end{equation}

Our SCRG strategy can be seamlessly integrated with existing on-policy sampling~\cite{agarwal2024policy} and off-policy sampling~\cite{ko2024distillm}. By incorporating an adaptive scheduler~\cite{ko2024distillm} for student-model generation, we enhance the efficiency of our sampling process.

\subsection{Token-level DAC-KL loss function}
\begin{figure}[!t]
  \centering
   \includegraphics[width=0.97\textwidth]{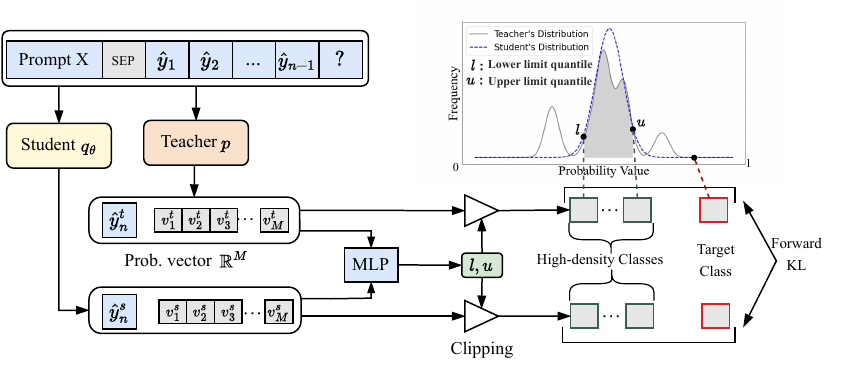} %
   \vspace{-0.2cm}
  \caption{The workflow of the DAC-KL loss function.}
  \label{DAC-KL}
\vspace{-0.5cm}
\end{figure}

The probability output of LLMs is a high-dimensional vector for each token. However, existing modified Kullback-Leibler divergence (KLD) loss functions, used as knowledge distillation objectives, struggle to effectively capture the valuable distribution with high semantic knowledge from the teacher network. They either underfitt the the teacher's distribution, as seen in forward KLD, or tend to overfit to a part of the high-probability region, as seen in reverse KLD. To address this issue, we design a Distribution-Adaptive Clipping Kullback-Leibler (DAC-KL) loss function (in Fig.~\ref{DAC-KL}) to capture high-density semantic regions of the teacher's output probability distribution, which can be more easily imitated by the student models with limited capacity.

The probability outputs at the $i^{th}$ token position of both the teacher and student models are high-dimensional probability vectors with $M$ tokens, which are denoted as 
\begin{equation}
\begin{aligned}
\hat{y}^t_i=p(.|y^s_{<i},x)=[v^t_1,v^t_2,...,v^t_M]\in\mathbb{R}^M,\\
\hat{y}^s_i=q_\theta(.|y^s_{<i},x)=[v^s_1,v^s_2,...,v^s_M]\in\mathbb{R}^M.
\end{aligned}
\end{equation}

We input these two probability vectors to a learnable MLP sub-network $f_{sub}$ to predict the upper limit quantile $u\in [0,1]$ and the lower limit quantile $l\in [0, u]$ of the probability distribution $\hat{y}^t_i$. We formulate this process as
\begin{equation}
\begin{aligned}
u, l = \sigma(f_{\text{sub}}(\hat{y}_i^t \mid sort(\hat{y}_i^t) \mid \hat{y}_i^s)),\\
\end{aligned}
\end{equation}
where $\sigma(\cdot )$ is the SIGMOID activation, $sort(\cdot )$ is the decending sort operation, and $\mid$ represents the concatenation operation, $l$ is clipped into the range $[0, u]$.

The predicted quantiles $u$ and $l$ are used to adaptively clip out the high-density semantic classes from the teacher's probability vector $\hat{y}^t_i$. We utilize the clipped high-density classes and the target class with the most probability value to construct a new probability vector $\hat{y}^{t*}_i$, which is formulated as
\begin{equation}
\begin{aligned}
\hat{y}^{t*}_i = \left[ \left\{ v^t_i \mid l \leq v^t_i \leq u \right\}_{1 \leq i \leq M} \mid \max(v^t_1, v^t_2, \ldots, v^t_M) \right].
\end{aligned}
\end{equation}

The high-density classes and the target class contain the most knowledge in the teacher's probability distribution.
Based on the corresponding positions of the clipped classes and target class of $\hat{y}^{t*}_i$, we construct the student's new probability vector $\hat{y}^{s*}_i$. Then, we adopt a vanilla KLD to calculate the sum of token-wise distillation loss and the final loss is calculated on the dataset ($X$,$Y$):
\begin{equation}
\begin{aligned}
\mathcal{L}_{\text{DAC-KLD}} =E_{x\sim X}[\tfrac{1}{L_{y^{s*}}}\sum_{i=1}^{L_{y^{s*}}}\hat{y}^{t*}_ilog\frac{\hat{y}^{t*}_i}{\hat{y}^{s*}_i}],
\end{aligned}
\end{equation}
where $L_{y^{s*}}$ is the length of the sequence generated from the proposed SCRG strategy.

\subsection{Span-level correlation consistency}
\begin{figure}[!t]
  \centering
   \includegraphics[width=\textwidth]{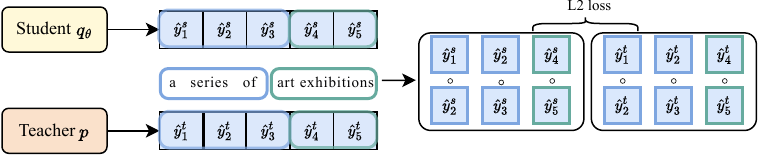} %
  \caption{The workflow of the span-level correlation distillation. $\circ$ denotes Hadamard multiplication.}
  \label{span}
  \vspace{-0.5cm}
\end{figure}

Motivated by the work~\cite{liu2022multi}, we utilize the pre-defined chunker~\cite{kiss2006unsupervised} to extract spans (including noun phrases, verb phrases, and prepositional phrases) that have complete meanings from the input sequences, which split a sequence into several token sets. For each token in the input sequence, LLMs predict a high-dimensional probability vector. The relations between tokens within the same span should maintain consistent relations in the transformed probability space. Constraining the relation consistency between the outputs of the student and the teacher models is crucial to transfer semantic knowledge, as shown in Fig.~\ref{span}.

We divide a probability sequence $\left [\hat{y}_1,\hat{y}_{2},...,\hat{y}_{n} \right ] $ into $n_s$ spans $s=\left [s_1, s_2,...,s_{n_s}\right ]$ accoreding to the pre-defined span priors from $\left [y_1,y_{2},...,y_{n} \right ] $. Here, $s_i = \left [\hat{y}_j,\hat{y}_{j+1},...,\hat{y}_{j+n_{s_i}-1} \right ] $ represents $i^{th} span$, whcih starts at the $j^{th}$ token of the sequence and contains $n_{s_i}$ tokens. 
Both the student and teacher model outputs adhere to the same span priors for token divisions. Consequently, we divide the probability outputs of the student and teacher models into spans, denoting the $i^{th}$ span as
\begin{equation}
\begin{aligned}
s^s_i = \left [\hat{y}^s_j,\hat{y}^s_{j+1},...,\hat{y}^s_{j+n_{s_i}-1}\right ], s^t_i = \left [\hat{y}^t_j,\hat{y}^t_{j+1},...,\hat{y}^t_{j+n_{s_i}-1}\right ].
\end{aligned}
\end{equation}

Next, we calculate the correlation between two adjacent tokens within the same spans and ensure consistency of this correlation between the probability outputs of the student model and the teacher model. To achieve this, we utilize the L2 distance to align the consistency. The span consistency loss is defined as follows:
\begin{equation}
\begin{aligned}
\mathcal{L}_{\text{span}} =E_{x\sim X}[\frac{1}{n_s} \sum_{i=1}^{n_s} \frac{1}{n_{s_i}}\sum_{(\hat{y}^s_{j},\hat{y}^s_{j+1})\in s^s_i,(\hat{y}^t_{j},\hat{y}^t_{j+1})\in s^t_i}^{} \left \| \hat{y}^s_{j}\circ\hat{y}^s_{j+1}-\hat{y}^t_{j}\circ\hat{y}^t_{j+1} \right \| _2],
\end{aligned}
\end{equation}

where $\left | \cdot \right |_2$ represents the L2 distance function, and $\circ$ denotes the Hadamard multiplication operation calculating correlation in the high-dimensional probability space. It is important to note that the output sequence is also generated by the student using the SCRG strategy. For simplicity, we adopt a standard symbol representation for $\hat{y}^t_{j}$ and $\hat{y}^s_{j}$ instead of $\hat{y}^{t*}_{j}$ and $\hat{y}^{s*}_{j}$.

\subsection{Overall Optimization}
We use the proposed KD method in the SFT stage of based models. The student model is supervised by the distillation loss, guided by the finetuned teacher model, and also supervised by the SFT loss. The overall optimization objective for the student model is formulated as
\begin{equation}
\begin{aligned}
\mathcal{L}_{overall} = \mathcal{L}_{\text{SFT}} + \mathcal{L}_{\text{DAC-KLD}} + \mathcal{L}_{\text{span}}.
\end{aligned}
\end{equation}
where $\mathcal{L}_{\text{SFT}}$ represents the SFT loss, $\mathcal{L}_{\text{DAC-KLD}}$ represents the distillation loss using the DAC-KLD object, and $\mathcal{L}_{\text{span}}$ represents the span consistency loss which assists the distillation process.

\section{Experiments}
In this section, we experiment by initially fine-tuning a large model on the dataset comprising instructions and corresponding responses $(X, Y)$, establishing it as the teacher model $p$. Subsequently, we examine various knowledge distillation methods for distilling a smaller student model under the guidance of the teacher, evaluating the instruction-following performance of the distilled model.

\subsection{Experimental description}

\noindent \textbf{Dataset and evaluation metrics.}
We conduct the KD experiments on five instruction-following datasets:
(1) Dolly Evaluation~\cite{dolly2023introducing} is a a sampled subset of atabricks-dolly-15k~\footnote{https://github.com/databrickslabs/dolly/tree/master} (Dolly) dataset consists of 500 samples. It covers various behavioural categories such as brainstorming, classification, closed QA, generation, information extraction, open QA, and summarization;
(2) Self-Instruct~\cite{wang2022self} is a dataset for language models' ability to understand and follow instructions.  It incorporates 252 expert-written tasks;
(3) Vicuna~\cite{wang2022super} is a dataset consisting of 80 challenging questions used for evaluating the Vicuna model. It follows the evaluation methodology introduced by MiniLLM~\cite{gu2023minillm};
(4) Super-Natural Instruction~\cite{wang2022super} is introduced as a benchmark, and this dataset contains 1,616 diverse NLP tasks along with their expert-written instructions. It covers 76 different task types, and its test set consists of 9K samples from 119 tasks;
(5) Unnatural Instruction~\cite{honovich2022unnatural} dataset comprises 240K instructions generated by AI with minimal human involvement. It shows that AI-generated data can be as effective as human-created data for training language models. The core component of this dataset has 60K samples.

We use the ROUGE-L~\cite{lin2004rouge} metric to evaluate the model-generated results and report the average scores of 5 generations for each prompt with different random seeds ($10$, $20$, $30$, $40$, $50$) for all test datasets. ROUGE-L evaluates the precision of the model's output by measuring the longest common subsequence between the generated text and the reference text. It is well-suited for large-scale instruction-following evaluation due to its ability to capture both sentence-level structure and content.

\noindent \textbf{Base models and baselines.}
We distil four kinds of teacher-student model pairs with different model sizes: LLAMA2~\cite{touvron2023llama} (13B teacher, 7B student), OpenLLAMA2~\cite{geng2023openllama} (7B teacher, 3B student), OPT~\cite{zhang2022opt} (6.7B teacher, 1.3B student), GPT2~\cite{radford2019language} (1.5B teacher, 0.1B student).

We benchmark our method against several advanced knowledge distillation methods:
(1) SFT Fine-tunes the student model on a fixed dataset in a vanilla manner;
(2) KD ~\cite{hinton2015distilling} utilizes KLD on a fixed dataset;
(3) SeqKD~\cite{kim2016sequence} fine-tunes on a teacher-generated dataset;
(4) ImitKD~\cite{lin2020autoregressive} utilizes KLD on a dataset generated by the student model;
(5) GKD~\cite{agarwal2024policy} utilizes Jensen-Shannon Divergence (JSD)~\cite{agarwal2024policy} on a mixture of a student-generated dataset and a fixed dataset;
(6) MiniLLM~\cite{gu2023minillm} utilizes a policy gradient approach on a dataset generated by the student model;
(7) DistiLLM~\cite{ko2024distillm} utilizes Skew KLD on a student-generated dataset sampling with an off-policy scheduler.

All of our baseline experiments are re-implemented using the open-source code~\footnote{https://github.com/jongwooko/distillm} on the same GPU servers utilized by our method. Additionally, we execute these experiments using the exact hyper-parameters as specified in the original codebase.

\noindent \textbf{Training details.}
We follow MiniLLM~\cite{gu2023minillm} to finetune base models using the training set of the databricks-dolly-15k. Dolly is divided into 14K samples as the training set and equally left 500 samples for validation and testing, respectively.
After the fine-tuning process, we select the best-performing model based on its validation set of the Dolly dataset. We then proceeded to test this selected model on the test sets of the five above-mentioned datasets. 

For training the teacher and student models, we utilize four A100 (40GB) GPUs for the OpenLLAMA2, OPT, and GPT2 models and four A800 (80GB) GPUs for the LLAMA2 models. A fixed learning rate of 5e-4 is applied consistently across all experiments. Specifically, for the LLAMA2, OpenLLAMA2, and OPT models, we follow DistiLLM~\cite{ko2024distillm}, employing low-rank adaptation (LoRA) for the query and value weights with a rank of 16 for 10 epochs. In contrast, for the GPT2 models, we fine-tune all parameters for 20 epochs.

\begin{table}[!ht] \fontsize{7}{8}\selectfont 
    \caption{Comparison of state-of-the-art knowledge distillation methods evaluated by the ROUGE-L metric~\cite{lin2004rouge}. `Average' is the average score on the five test datasets  The bold and underlined markings signify the best and second-best results, respectively.} 
	\begin{center}
	\renewcommand\tabcolsep{4pt}
	\begin{threeparttable}
	\begin{tabular}{cc|c|ccccc|c}
		\toprule[1.2pt]
		\multicolumn{2}{c|}{\multirow{2}{*}{Methods}}& \multicolumn{1}{c|}{\multirow{2}{*}{Parameters}} & \multicolumn{5}{c}{Datasets}   \\ 
            \cmidrule(r){4-9}
            ~& ~& ~& Dolly Evaluation& Self-Instruct & Vicuna  &Super-Natural& Unnatural &Average \\
		\midrule
		\multicolumn{1}{c|}{\multirow{10}{*}{LLAMA2}}  & Teacher (SFT) & 13B & 29.8241 &21.0617 &19.4909 & 35.8318& 35.7802&28.3978
   \\
		\cmidrule(r){2-9}
		\multicolumn{1}{c|}{~} & SFT &\multirow{7}{*}{7B}  & 27.3504 & 28.4430 & 18.7567 & 28.4430 & 30.2788&26.6544 
 \\
          \multicolumn{1}{c|}{~}  &KD~\cite{hinton2015distilling} &~    & 27.0737 & 20.7076 & 17.9850 & 30.3350 & 31.4926&25.5188
 \\
          \multicolumn{1}{c|}{~}  &SeqKD~\cite{kim2016sequence} &~   & 26.2689 & 19.0278 & 18.4602 & 25.9461 & 28.1010&23.5608
 \\
          \multicolumn{1}{c|}{~}  &ImitKD~\cite{lin2020autoregressive} &~   & 27.4359 & 20.6792 & 18.8058 & 29.1726 & 30.5764 &25.3340
\\
          \multicolumn{1}{c|}{~}  &GKD~\cite{agarwal2024policy} &~   & 28.4662 & 22.1717 & 20.7564 & 33.3325 & 33.2682 &27.5990
 \\
          \multicolumn{1}{c|}{~}  &MiniLLM~\cite{gu2023minillm} &~   & 30.6447 & 23.9493 & \textbf{22.3010} & 34.3454 & 36.0828 &29.4646
 \\
          \multicolumn{1}{c|}{~}  &DistiLLM~\cite{ko2024distillm} &~  & \uline{30.7277} & \uline{25.2181} & 20.8356 & \uline{36.1154} & \uline{37.5072} & \uline{30.0808}
\\
          \multicolumn{1}{c|}{~}  &Ours &~  & \cellcolor{gray!20}\textbf{31.9195} & \cellcolor{gray!20}\textbf{25.4937} & \cellcolor{gray!20}\uline{21.7810} & \cellcolor{gray!20}\textbf{37.9154} & \cellcolor{gray!20}\textbf{38.1257} &\cellcolor{gray!20}\textbf{31.0471}
\\
          \cmidrule(r){1-9}
          \cmidrule(r){1-9}
\multicolumn{1}{c|}{\multirow{10}{*}{OpenLLAMA2}}  & Teacher (SFT) & 7B & 27.5100 & 17.9400 & 17.6900 & 32.7500 & 31.4000&25.4580
 \\
		\cmidrule(r){2-9}
		\multicolumn{1}{c|}{~} & SFT &\multirow{7}{*}{3B}  & 24.4000 & 16.1300 & 16.5600 & 27.4862 & 28.0500&22.5252
 \\
          \multicolumn{1}{c|}{~}  &KD~\cite{hinton2015distilling} &~    & 25.4814 & 19.1805 & 16.6562 & 31.3307 & 31.8136&24.8924
 \\
          \multicolumn{1}{c|}{~}  &SeqKD~\cite{kim2016sequence} &~ & 24.8184 & 16.0980 & 17.2718 & 29.4081 & 28.7395&23.2672
 \\
          \multicolumn{1}{c|}{~}  &ImitKD~\cite{lin2020autoregressive} &~  & 25.3600 & 18.1600 & 17.5700 & 31.0900 & 28.9600 &24.2280
 \\
          \multicolumn{1}{c|}{~}  &GKD~\cite{agarwal2024policy} &~  & 26.8525 & 20.1060 & 18.4337 & 34.4383 & 32.4797 &26.4621
\\
          \multicolumn{1}{c|}{~}  &MiniLLM~\cite{gu2023minillm} &~  & \uline{28.4950} & \textbf{21.7770} & \textbf{20.6260} & \uline{35.4001} & 34.7011 &\uline{28.1999}
\\
          \multicolumn{1}{c|}{~}  &DistiLLM~\cite{ko2024distillm} &~  & 27.8546 & 19.3456 & \uline{19.1723} & 34.4973 & \uline{34.9434} &27.1627
\\
          \multicolumn{1}{c|}{~}  &Ours &~  & \cellcolor{gray!20}\textbf{29.3062} & \cellcolor{gray!20}\uline{20.5835} & \cellcolor{gray!20}19.0086 & \cellcolor{gray!20}\textbf{37.6171} & \cellcolor{gray!20}\textbf{37.2410} &\cellcolor{gray!20}\textbf{28.8724}
 \\
          \cmidrule(r){1-9}
          \cmidrule(r){1-9}
        \multicolumn{1}{c|}{\multirow{10}{*}{OPT}}  & Teacher (SFT) & 6.7B & 25.8758 & 14.8408 & 16.4199 & 24.9551 & 25.8377&21.5859
 \\
		\cmidrule(r){2-9}
		\multicolumn{1}{c|}{~} & SFT &\multirow{7}{*}{1.3B}  & 22.7595 & 11.9784 & 15.2267 & 22.8556 & 24.5763 &19.4793
 \\
          \multicolumn{1}{c|}{~}  &KD~\cite{hinton2015distilling} &~ & 22.4476 & 13.4676 & 13.9975 & 23.7679 & 25.4132 &19.8188
 \\
          \multicolumn{1}{c|}{~}  &SeqKD~\cite{kim2016sequence} &~ & 22.4556 & 12.1588 & 14.8157 & 21.4574 & 24.5907 &19.0956
 \\
          \multicolumn{1}{c|}{~}  &ImitKD~\cite{lin2020autoregressive} &~ & 21.6624 & 12.9286 & 15.8039 & 22.0426 & 24.9619 &19.4799
\\
          \multicolumn{1}{c|}{~}  &GKD~\cite{agarwal2024policy} &~ & 22.5062 & 12.8309 & 15.3303 & 23.8537 & 26.6441 &20.2330
\\
          \multicolumn{1}{c|}{~}  &MiniLLM~\cite{gu2023minillm} &~  & 24.3168 & 13.5880 & \textbf{17.4633} & 26.6789 & 28.7968 &22.1688
 \\
          \multicolumn{1}{c|}{~}  &DistiLLM~\cite{ko2024distillm} &~  & \uline{24.7311} & \uline{14.9932} & \uline{16.3293} & \uline{27.1037} & \uline{29.3285} &\uline{22.4972}
\\
          \multicolumn{1}{c|}{~}  &Ours &~  & \cellcolor{gray!20}\textbf{27.1486} & \cellcolor{gray!20}\textbf{17.3016} & \cellcolor{gray!20}14.8491 & \cellcolor{gray!20}\textbf{32.0618} & \cellcolor{gray!20}\textbf{34.9709} &\cellcolor{gray!20}\textbf{25.2664}
 \\
		\cmidrule(r){1-9}
  		\cmidrule(r){1-9}
        \multicolumn{1}{c|}{\multirow{10}{*}{GPT2}}  & Teacher (SFT) & 1.5B & 27.0357 & 14.5594 & 16.7390 & 24.9659 & 29.4874 &22.5575
\\
        \cmidrule(r){2-9}
		\multicolumn{1}{c|}{~} & SFT &\multirow{7}{*}{0.1B}  & 23.8269 & 9.6682 & 14.9022 & 16.4117 & 18.3221 &16.6262
\\
          \multicolumn{1}{c|}{~}  &KD~\cite{hinton2015distilling} &~    & 23.2172 & 10.0899 & 14.9954 & 15.4826 & 18.9597 &16.5490
 \\
          \multicolumn{1}{c|}{~}  &SeqKD~\cite{kim2016sequence} &~  & 23.7248 & 10.3935 & 14.6558 & 19.8119 & 22.7425 &18.2657
 \\
          \multicolumn{1}{c|}{~}  &ImitKD~\cite{lin2020autoregressive} &~  & 21.7724 & 10.1876 & 15.4640 & 17.1918 & 20.8907 &17.1013
 \\
          \multicolumn{1}{c|}{~}  &GKD~\cite{agarwal2024policy} &~ & 23.3150 & 10.3364 & 15.9384 & 16.0802 & 17.7699 &16.6880
\\
          \multicolumn{1}{c|}{~}  &MiniLLM~\cite{gu2023minillm} &~ & 23.8142 & 12.2771 & \uline{17.0158} & 23.8555 & 24.9101 &20.3745
 \\
          \multicolumn{1}{c|}{~}  &DistiLLM~\cite{ko2024distillm} &~ & \uline{25.6114} & \uline{12.5988} & 16.7521 & \uline{24.6374} & \textbf{27.5827} &\uline{21.4365}
\\
          \multicolumn{1}{c|}{~}  &Ours &~  & \cellcolor{gray!20}\textbf{26.5614} & \cellcolor{gray!20}\textbf{13.1174} & \cellcolor{gray!20}\textbf{17.6781} & \cellcolor{gray!20}\textbf{24.6973} & \cellcolor{gray!20}\uline{27.4025} &\cellcolor{gray!20}\textbf{21.8913}
\\
	\bottomrule[1.2pt] 
	\end{tabular}
	\end{threeparttable}
 	\end{center}
	\label{exp_rouge}
 \vspace{-0.7cm}
\end{table}

\subsection{Comparison with state-of-the-art KD methods}

We present the quantitative comparison of state-of-the-art knowledge distillation methods evaluated using the ROUGE-L metric in Table~\ref{exp_rouge}. It is observed that:

(1) Our method outperforms existing methods in most distillation tasks, with only a few achieving second-best results, across five test datasets, including the LLAMA2, OPT, OpenLLAMA2, and GPT2 series of large language models. Particularly for the OPT datasets, our method shows an average score improvement of over 12\% compared to the second-best performing methods.

(2) The KD methods, such as GKD, MiniLLM, and DistiLLM, utilizing student-generated datasets show a greater improvement in enhancing student performance compared to those using the fixed dataset. Furthermore, the distilled student models generally outperform the teacher models, which can be attributed to the mismatch between teacher-forcing training and free-run generation, i.e., exposure bias~\cite{bengio2015scheduled}. Our method can improve the performance of all student models on average scores of the five test datasets by at least 15\%.

(3) We also provide some representative instruction-following cases in Section~\ref{case}, further highlighting the effectiveness and superiority of our method in achieving high-quality answers.

\subsection{Ablation analysis}
We conduct an ablation analysis of the proposed methods on the Dolly Validation set, Dolly Evaluation set and Self-Instruct dataset.

\noindent \textbf{Overall Ablation.}
We conduct an overall ablation study to validate the effectiveness of the proposed multi-granularity semantic revision, in Table~\ref{overall_ablation}.
Initially, employing sequence correction alone yields moderate performance improvement across all evaluation datasets compared to the vanilla result. Upon the addition of DAC-KL, an improvement is observed. A further enhancement is achieved with the inclusion of span-level relation distillation, resulting in more notable performance gains. The most significant improvement is witnessed when all components of the proposed method are combined, leading to the highest performance metrics across all evaluation datasets. This demonstrates that each component contributes to the overall enhancement of model performance, with the combined approach yielding the most substantial improvements.

\noindent \textbf{Different student-generation methods.}
To validate the effectiveness of the proposed SCRG strategy, we compare it with different student-generation methods for sampling the distillation dataset. As illustrated in Table~\ref{scrg_a}, we observe substantial performance enhancements with SCRG compared to existing student-generation methods. For on-policy sampling, We follow GKD~\cite{agarwal2024policy} to utilize a mixture of student-generated and fixed datasets. For off-policy sampling, we follow Distillm~\cite{ko2024distillm} to adopt an adaptive student-generation schedule for improved sample efficiency. Remarkably, when employing both off-policy and on-policy sampling methods, SCRG achieves notably higher scores across all evaluation metrics. This underscores the effectiveness of SCRG in augmenting performance by improving the quality and diversity of generated sequences. Additionally, we provide an example of SCRG in Section~\ref{example_SCRG}.

\noindent \textbf{Different distillation loss functions.}
To validate the effectiveness of the proposed DAC-KL loss, we compare it with different loss functions in Table~\ref{scrg_c}. The results demonstrate that DAC-KL significantly outperforms other loss functions across all evaluation metrics. This indicates that DAC-KL effectively captures high-density semantic regions of the teacher’s output probability distribution, facilitating easier imitation by the student models. Additionally, we provide visualized examples of the DAC-KL impact on the probability distribution of the teacher’s output depicted using kernel density estimation in Section~\ref{DAC_vis}.

\noindent \textbf{Different components involved in DAC-KL.} 
The DAC-KL loss guides the distillation process to effectively transfer knowledge from the high-density semantic classes and the target class of the teacher's probability outputs. As evidenced by the results in Table~\ref{scrg_b}, when both high-density and target classes are considered, the DAC-KL loss achieves the highest validation, evaluation, and self-instruct scores compared to other configurations. This indicates that focusing on these specific classes leads to better performance in knowledge distillation, highlighting the importance of targeting relevant semantic regions for the effective transfer of knowledge.

\begin{table}[!t]\fontsize{7}{8}\selectfont
\centering
\caption{Ablation study of the proposed multi-granularity semantic revision.}
\vspace{-0.2cm}
	\begin{center}
	\renewcommand\tabcolsep{6pt}
 \begin{threeparttable}
\label{tab:self-correcting}
\begin{tabular}{c|c|c|ccc}
    \toprule[1.2pt]
    Sequence-correcting & DAC-KL & Span Relation & Dolly Validation & Dolly Evaluation & Self-Instruct \\
    \midrule
    \XSolidBrush& \XSolidBrush&\XSolidBrush & 29.1874 & 24.1603 & 14.8578 \\
    \Checkmark  &\XSolidBrush &\XSolidBrush & 29.6982 & 24.5307 & 14.9485 \\
    \Checkmark& \Checkmark  & \XSolidBrush & 30.3486 & 26.9012  & 17.2392  \\
   \Checkmark  & \Checkmark & \Checkmark  & \textbf{31.2575} & \textbf{27.1486} & \textbf{17.3016} \\
    \bottomrule[1.2pt]
\end{tabular}    
\end{threeparttable}
\end{center}
 \label{overall_ablation}
 \vspace{-0.6cm}
\end{table}

\begin{table}[!t]
 \caption{Ablation studies on the proposed SCRG strategy and the DAC-KL loss.}
 \vspace{-0.2cm}
  \begin{subtable}{0.5\textwidth}
 \caption{Different student-generation methods}
    \centering
      \fontsize{7}{8}\selectfont
      \renewcommand\tabcolsep{2pt}
      \setlength{\abovecaptionskip}{0.1pt}
    \begin{threeparttable}
    \begin{tabular}{c|ccc}
    \toprule[1.2pt]
    Generation & Validation &  Evaluation & Self-Instruct \\
    \midrule
    On-policy~\cite{lin2020autoregressive} & 30.3786 & 26.0948 & 16.1853 \\
    Mixed~\cite{agarwal2024policy}  & 30.8335 & 26.4667 & 16.7789 \\
   Off-policy~\cite{ko2024distillm} & 30.4539 & 27.0961 & 16.7745 \\
   \midrule
    SCRG w.  off policy &31.0444  &27.1453 & 17.2574  \\
    SCRG w.  on policy & \textbf{31.2575} & \textbf{27.1486} & \textbf{17.3016} \\
    \bottomrule[1.2pt]
    \end{tabular} 
    \label{scrg_a}
    \end{threeparttable}
    \vspace{-0.1cm}
 \caption{Components involved in DAC-KL losses}
     \begin{tabular}{cc|ccc}
    \toprule[1.2pt]
    High-density &Target & Validation &  Evaluation & Self-Instruct \\
    \XSolidBrush & \Checkmark&29.3490  &24.3130 & 14.3810  \\
    \Checkmark & \XSolidBrush&21.3936  &19.5050 & 11.5035  \\
    \Checkmark & \Checkmark& \textbf{31.2575} & \textbf{27.1486} & \textbf{17.3016} \\
    \bottomrule[1.2pt]
    \end{tabular} 
    \label{scrg_b}
  \end{subtable}\hfill
  \begin{subtable}{0.5\textwidth}
   \vspace{4mm}
 \caption{Different distillation loss functions}
    \centering
      \fontsize{7}{10}\selectfont
      \renewcommand\tabcolsep{2pt}
      \setlength{\abovecaptionskip}{0.1pt}
    \begin{threeparttable}
    \begin{tabular}{c|ccc}
    \toprule[1.2pt]
    Loss Function & Validation &  Evaluation & Self-Instruct \\
    \midrule
    Forward KL & 28.9631 & 24.1922 & 14.5108 \\
    Reverse KL & 30.0209 & 25.6688 & 14.7184 \\
    Symmetric KL &30.2584  &27.0961  &16.7745  \\
    Generalized JSD~\cite{agarwal2024policy} &27.8759 &23.3144 & 14.3154 \\
    TVD~\cite{wen2023f} & 30.1973 & 25.0033 & 14.6138 \\
    SRKL~\cite{ko2024distillm} & 29.9858 & 25.4849 & 14.9514 \\
    SFKL~\cite{ko2024distillm} & 29.1226 & 25.1400 & 14.4412 \\
    \midrule
    DAC-KL  & \textbf{31.2575} & \textbf{27.14864} & \textbf{17.3016} \\
    \bottomrule[1.2pt]
    \end{tabular}     
    \end{threeparttable}
    \label{scrg_c}
  \end{subtable}
  \vspace{-0.7cm}
\end{table}

\section{Conclusion and Limitation}
In this paper, we address the challenges in knowledge distillation for LLMs by proposing a novel multi-level semantic revision approach at the sequence, token, and span levels. At the sequence level, our sequence correction and re-generation strategy improves reliability and diversity in student-generated sequences. At the token level, the DAC-KL loss function targets semantically salient regions in the teacher's probability distribution, filtering out redundant information. At the span level, input span priors ensure consistent transfer of semantic information across related tokens. Our experiments with four various model series, demonstrate the effectiveness of our approach, significantly improving student model performance over existing KD methods. In addition, our experiments and evaluations were conducted primarily on language models in specific domains. The effectiveness of our approach in other domains or tasks may vary, and further research is needed to explore its generalizability.

{
\small
\normalem
\bibliographystyle{unsrt}
\bibliography{reference}

\begin{thebibliography}{10}

\bibitem{kaplan2020scaling}
Jared Kaplan, Sam McCandlish, Tom Henighan, Tom~B Brown, Benjamin Chess, Rewon
  Child, Scott Gray, Alec Radford, Jeffrey Wu, and Dario Amodei.
\newblock Scaling laws for neural language models.
\newblock {\em arXiv preprint arXiv:2001.08361}, 2020.

\bibitem{Wei_2022}
Jason Wei, Yi~Tay, Rishi Bommasani, Colin Raffel, Barret Zoph, Sebastian
  Borgeaud, Dani Yogatama, Maarten Bosma, Denny Zhou, Donald Metzler, EdH. Chi,
  Tatsunori Hashimoto, Oriol Vinyals, Percy Liang, Jeff Dean, and William
  Fedus.
\newblock Emergent abilities of large language models.
\newblock Jun 2022.

\bibitem{Radford_Wu_Child_Luan_Amodei_Sutskever}
Alec Radford, Jeffrey Wu, Rewon Child, David Luan, Dario Amodei, and Ilya
  Sutskever.
\newblock Language models are unsupervised multitask learners.

\bibitem{Zhangetal}
Susan Zhang, Stephen Roller, Naman Goyal, Mikel Artetxe, Moya Chen, Shuohui
  Chen, Christopher Dewan, Mona Diab, Xian Li, Victoria Lin, Todor Mihaylov,
  Myle Ott, Sam Shleifer, Kurt Shuster, Daniel Simig, Singh Koura, Anjali
  Sridhar, Tianlu Wang, and Luke Zettlemoyer.
\newblock Opt: Open pre-trained transformer language models.

\bibitem{Brown_Mann}
TomB. Brown, Benjamin Mann, Nick Ryder, Melanie Subbiah, Jared Kaplan, Prafulla
  Dhariwal, Arvind Neelakantan, Pranav Shyam, Girish Sastry, Amanda Askell,
  Sandhini Agarwal, Ariel Herbert-Voss, Gretchen Krueger, Thomas Henighan,
  Rewon Child, Aditya Ramesh, DanielM. Ziegler, Jeffrey Wu, Clemens Winter,
  Christopher Hesse, Mark Chen, Eric Sigler, Mateusz Litwin, Scott Gray,
  Benjamin Chess, Jack Clark, Christopher Berner, Samuel McCandlish, Alec
  Radford, Ilya Sutskever, and Dario Amodei.
\newblock Language models are few-shot learners.
\newblock {\em arXiv: Computation and Language,arXiv: Computation and
  Language}, May 2020.

\bibitem{Touvron}
Hugo Touvron, Thibaut Lavril, Gautier Izacard, Xavier Martinet, Marie-Anne
  Lachaux, Timoth’ee Lacroix, Baptiste Rozi`ere, Naman Goyal, Eric Hambro,
  Faisal Azhar, Aurelien Rodriguez, Armand Joulin, Edouard Grave, and Guillaume
  Lample.
\newblock Llama: Open and efficient foundation language models.

\bibitem{zhang2022opt}
Susan Zhang, Stephen Roller, Naman Goyal, Mikel Artetxe, Moya Chen, Shuohui
  Chen, Christopher Dewan, Mona Diab, Xian Li, Xi~Victoria Lin, et~al.
\newblock Opt: Open pre-trained transformer language models.
\newblock {\em arXiv preprint arXiv:2205.01068}, 2022.

\bibitem{wang2022self}
Yizhong Wang, Yeganeh Kordi, Swaroop Mishra, Alisa Liu, Noah~A Smith, Daniel
  Khashabi, and Hannaneh Hajishirzi.
\newblock Self-instruct: Aligning language models with self-generated
  instructions.
\newblock {\em arXiv preprint arXiv:2212.10560}, 2022.

\bibitem{hinton2015distilling}
Geoffrey Hinton, Oriol Vinyals, and Jeff Dean.
\newblock Distilling the knowledge in a neural network.
\newblock {\em arXiv preprint arXiv:1503.02531}, 2015.

\bibitem{agarwal2024policy}
Rishabh Agarwal, Nino Vieillard, Yongchao Zhou, Piotr Stanczyk, Sabela~Ramos
  Garea, Matthieu Geist, and Olivier Bachem.
\newblock On-policy distillation of language models: Learning from
  self-generated mistakes.
\newblock In {\em The Twelfth International Conference on Learning
  Representations}, 2024.

\bibitem{gu2023minillm}
Yuxian Gu, Li~Dong, Furu Wei, and Minlie Huang.
\newblock Minillm: Knowledge distillation of large language models.
\newblock In {\em The Twelfth International Conference on Learning
  Representations}, 2023.

\bibitem{ko2024distillm}
Jongwoo Ko, Sungnyun Kim, Tianyi Chen, and Se-Young Yun.
\newblock Distillm: Towards streamlined distillation for large language models.
\newblock {\em arXiv preprint arXiv:2402.03898}, 2024.

\bibitem{kim2016sequence}
Yoon Kim and Alexander~M Rush.
\newblock Sequence-level knowledge distillation.
\newblock {\em arXiv preprint arXiv:1606.07947}, 2016.

\bibitem{jiao2019tinybert}
Xiaoqi Jiao, Yichun Yin, Lifeng Shang, Xin Jiang, Xiao Chen, Linlin Li, Fang
  Wang, and Qun Liu.
\newblock Tinybert: Distilling bert for natural language understanding.
\newblock {\em arXiv preprint arXiv:1909.10351}, 2019.

\bibitem{adriana2015fitnets}
Romero Adriana, Ballas Nicolas, K~Samira Ebrahimi, Chassang Antoine, Gatta
  Carlo, and Bengio Yoshua.
\newblock Fitnets: Hints for thin deep nets.
\newblock {\em Proc. ICLR}, 2(3):1, 2015.

\bibitem{liang2020mixkd}
Kevin~J Liang, Weituo Hao, Dinghan Shen, Yufan Zhou, Weizhu Chen, Changyou
  Chen, and Lawrence Carin.
\newblock Mixkd: Towards efficient distillation of large-scale language models.
\newblock {\em arXiv preprint arXiv:2011.00593}, 2020.

\bibitem{sanh2019distilbert}
Victor Sanh, Lysandre Debut, Julien Chaumond, and Thomas Wolf.
\newblock Distilbert, a distilled version of bert: smaller, faster, cheaper and
  lighter.
\newblock {\em arXiv preprint arXiv:1910.01108}, 2019.

\bibitem{liang2023less}
Chen Liang, Simiao Zuo, Qingru Zhang, Pengcheng He, Weizhu Chen, and Tuo Zhao.
\newblock Less is more: Task-aware layer-wise distillation for language model
  compression.
\newblock In {\em International Conference on Machine Learning}, pages
  20852--20867. PMLR, 2023.

\bibitem{sun2019patient}
Siqi Sun, Yu~Cheng, Zhe Gan, and Jingjing Liu.
\newblock Patient knowledge distillation for bert model compression.
\newblock {\em arXiv preprint arXiv:1908.09355}, 2019.

\bibitem{liu2022multi}
Chang Liu, Chongyang Tao, Jiazhan Feng, and Dongyan Zhao.
\newblock Multi-granularity structural knowledge distillation for language
  model compression.
\newblock In {\em Proceedings of the 60th Annual Meeting of the Association for
  Computational Linguistics (Volume 1: Long Papers)}, pages 1001--1011, 2022.

\bibitem{ouyang2022training}
Long Ouyang, Jeffrey Wu, Xu~Jiang, Diogo Almeida, Carroll Wainwright, Pamela
  Mishkin, Chong Zhang, Sandhini Agarwal, Katarina Slama, Alex Ray, et~al.
\newblock Training language models to follow instructions with human feedback.
\newblock {\em Advances in neural information processing systems},
  35:27730--27744, 2022.

\bibitem{achiam2023gpt}
Josh Achiam, Steven Adler, Sandhini Agarwal, Lama Ahmad, Ilge Akkaya,
  Florencia~Leoni Aleman, Diogo Almeida, Janko Altenschmidt, Sam Altman,
  Shyamal Anadkat, et~al.
\newblock Gpt-4 technical report.
\newblock {\em arXiv preprint arXiv:2303.08774}, 2023.

\bibitem{chen2024knowledge}
Hongzhan Chen, Xiaojun Quan, Hehong Chen, Ming Yan, and Ji~Zhang.
\newblock Knowledge distillation for closed-source language models.
\newblock {\em arXiv preprint arXiv:2401.07013}, 2024.

\bibitem{jiang2023lion}
Yuxin Jiang, Chunkit Chan, Mingyang Chen, and Wei Wang.
\newblock Lion: Adversarial distillation of closed-source large language model.
\newblock {\em arXiv preprint arXiv:2305.12870}, 2023.

\bibitem{hsieh2023distilling}
Cheng-Yu Hsieh, Chun-Liang Li, Chih-Kuan Yeh, Hootan Nakhost, Yasuhisa Fujii,
  Alexander Ratner, Ranjay Krishna, Chen-Yu Lee, and Tomas Pfister.
\newblock Distilling step-by-step! outperforming larger language models with
  less training data and smaller model sizes.
\newblock {\em arXiv preprint arXiv:2305.02301}, 2023.

\bibitem{taori2023stanford}
Rohan Taori, Ishaan Gulrajani, Tianyi Zhang, Yann Dubois, Xuechen Li, Carlos
  Guestrin, Percy Liang, and Tatsunori~B Hashimoto.
\newblock Stanford alpaca: An instruction-following llama model, 2023.

\bibitem{chiang2023vicuna}
Wei-Lin Chiang, Zhuohan Li, Zi~Lin, Ying Sheng, Zhanghao Wu, Hao Zhang, Lianmin
  Zheng, Siyuan Zhuang, Yonghao Zhuang, Joseph~E Gonzalez, et~al.
\newblock Vicuna: An open-source chatbot impressing gpt-4 with 90\%* chatgpt
  quality.
\newblock {\em See https://vicuna. lmsys. org (accessed 14 April 2023)},
  2(3):6, 2023.

\bibitem{peng2023instruction}
Baolin Peng, Chunyuan Li, Pengcheng He, Michel Galley, and Jianfeng Gao.
\newblock Instruction tuning with gpt-4.
\newblock {\em arXiv preprint arXiv:2304.03277}, 2023.

\bibitem{kiss2006unsupervised}
Tibor Kiss and Jan Strunk.
\newblock Unsupervised multilingual sentence boundary detection.
\newblock {\em Computational linguistics}, 32(4):485--525, 2006.

\bibitem{dolly2023introducing}
Free Dolly.
\newblock Introducing the world’s first truly open instruction-tuned llm.
  databricks. com, 2023.

\bibitem{wang2022super}
Yizhong Wang, Swaroop Mishra, Pegah Alipoormolabashi, Yeganeh Kordi, Amirreza
  Mirzaei, Anjana Arunkumar, Arjun Ashok, Arut~Selvan Dhanasekaran, Atharva
  Naik, David Stap, et~al.
\newblock Super-naturalinstructions: Generalization via declarative
  instructions on 1600+ nlp tasks.
\newblock {\em arXiv preprint arXiv:2204.07705}, 2022.

\bibitem{honovich2022unnatural}
Or~Honovich, Thomas Scialom, Omer Levy, and Timo Schick.
\newblock Unnatural instructions: Tuning language models with (almost) no human
  labor.
\newblock {\em arXiv preprint arXiv:2212.09689}, 2022.

\bibitem{lin2004rouge}
Chin-Yew Lin.
\newblock Rouge: A package for automatic evaluation of summaries.
\newblock In {\em Text summarization branches out}, pages 74--81, 2004.

\bibitem{touvron2023llama}
Hugo Touvron, Louis Martin, Kevin Stone, Peter Albert, Amjad Almahairi, Yasmine
  Babaei, Nikolay Bashlykov, Soumya Batra, Prajjwal Bhargava, Shruti Bhosale,
  et~al.
\newblock Llama 2: Open foundation and fine-tuned chat models.
\newblock {\em arXiv preprint arXiv:2307.09288}, 2023.

\bibitem{geng2023openllama}
Xinyang Geng and Hao Liu.
\newblock Openllama: An open reproduction of llama.
\newblock {\em URL: https://github. com/openlm-research/open\_llama}, 2023.

\bibitem{radford2019language}
Alec Radford, Jeffrey Wu, Rewon Child, David Luan, Dario Amodei, Ilya
  Sutskever, et~al.
\newblock Language models are unsupervised multitask learners.
\newblock {\em OpenAI blog}, 1(8):9, 2019.

\bibitem{lin2020autoregressive}
Alexander Lin, Jeremy Wohlwend, Howard Chen, and Tao Lei.
\newblock Autoregressive knowledge distillation through imitation learning.
\newblock {\em arXiv preprint arXiv:2009.07253}, 2020.

\bibitem{bengio2015scheduled}
Samy Bengio, Oriol Vinyals, Navdeep Jaitly, and Noam Shazeer.
\newblock Scheduled sampling for sequence prediction with recurrent neural
  networks.
\newblock {\em Advances in neural information processing systems}, 28, 2015.

\bibitem{wen2023f}
Yuqiao Wen, Zichao Li, Wenyu Du, and Lili Mou.
\newblock f-divergence minimization for sequence-level knowledge distillation.
\newblock {\em arXiv preprint arXiv:2307.15190}, 2023.

\end{thebibliography}
}

\newpage
\appendix
\section{Appendix / supplemental material}

\subsection{Social Impact}
The primary objective of this study is to contribute to the advancement of the field of Machine Learning, without explicitly emphasizing any specific societal consequences. Although smaller models can lead to positive outcomes, such as reduced emissions, it is crucial to conduct a comprehensive study on potential biases associated with model compression. However, there are potential negative impacts to consider. Model compression may inadvertently exacerbate existing biases within data, leading to unfair outcomes, particularly for underrepresented groups. Additionally, the simplification involved in compression could result in the loss of critical nuances and reduce the model's ability to handle complex tasks accurately.

\subsection{Visualized probability distribution of the teacher model}
  \label{DAC_vis}
  
\begin{figure}[!h]
  \centering
   \includegraphics[width=\textwidth]{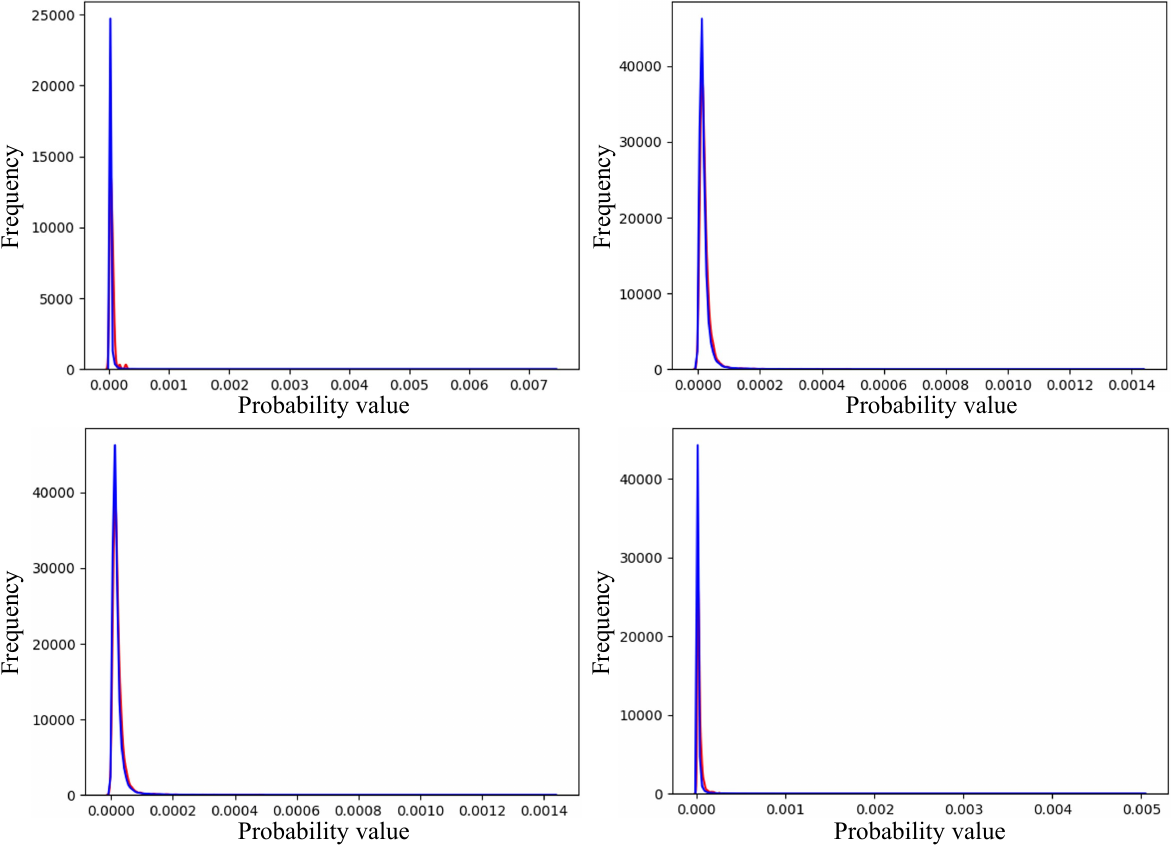} %
  \caption{
Examples of the probability distribution of the teacher's output are depicted using kernel density estimation. The original distribution is represented by the \textcolor{blue}{blue} line, while the distribution of the adaptively clipped probability classes is shown by the \textcolor{red}{red} line. 
From this picture, we can observe that the DAC-KL loss constrains the regions of probability distribution with dense semantic knowledge. Enforcing student model to  imitate the distribution of these regions can effectively mitigate the training interference caused by low-semantic regions for student models with limited learning capacity.}

  \vspace{-0.4cm}
\end{figure}

\newpage
\subsection{Examples of model responses on the instruction-following task.}
 \label{case}

\begin{table}[!ht]
  \caption{Instruction-following cases from the databricks-dolly-15k dataset.}
  \centering
  \begin{tabularx}{\textwidth}{@{}l|X@{}}
    \toprule[1.2pt]
    \multicolumn{2}{c}{Case \#1, Category: classification} \\
    \midrule
    Instruction &If we were playing a game where we had to identify things that can be found inside a house, which of these would we call out: car, chair, table, park, cloud, microwave.  \\
    \midrule
    Input & - \\
    \midrule
    Groundtruth &Inside a house you can find a chair, a table and a microwave. Cars, clouds and parks are found outside of houses.  \\
    \midrule
    \midrule
    SFT & car, chair, table, microwave.\\
    \midrule
    KD & Car, chair, table and microwave are things that can be found in a house. Park and cloud are unlikely to be found in a house but are things that can be seen from a house.\\
    \midrule
    SeqKD &We would call out: chair, table, car and microwave.\\
        \midrule
    ImitKD &Car, chair, table and microwave are items that can be found inside of a house. Park and cloud are outdoors items. \\
    \midrule
    MiniLLM &Car, chair, table, and microwave can all be found inside a house. Park and cloud cannot be found inside a house because they are neither objects nor locations.\\
    \midrule
    GKD & You can find a car, chair, table and microwave inside a house. A park and cloud are not found inside a house.\\
    \midrule
    DISTILLM & In this list, we can find a car, chair, table and microwave. A park and a cloud are not things you would typically find inside a house.\\
    \midrule
    Ours &In this game, a car, park, and cloud would not be things that could be found inside a house. A chair, table, and microwave would be things that could be find inside a house. \\
    \bottomrule[1.2pt] 
  \end{tabularx}
  \label{case1}
\end{table}

\begin{table}[!ht]
  \caption{Instruction-following cases from the databricks-dolly-15k dataset.}
  \centering
  \begin{tabularx}{\textwidth}{@{}l|X@{}}
    \toprule[1.2pt]
    \multicolumn{2}{c}{Case \#2, Category: closed QA} \\
    \midrule
    Instruction &How many managers have Arsenal had?  \\
    \midrule
    Input & There have been twenty permanent and eight caretaker managers of Arsenal since 1897; Stewart Houston has managed the club in two separate spells as caretaker. The most successful person to manage Arsenal is Ars\'e8ne Wenger, who won three Premier League titles, seven FA Cups and seven Community Shields between 1996 and 2018. Wenger is the club's longest-serving manager; he surpassed George Allison's record of 13 years in October 2009. Two Arsenal managers have died in the job \u2013 Herbert Chapman and Tom Whittaker.\\
    \midrule
    Groundtruth &Arsenal have had twenty permanent and eight caretake managers. \\
    \midrule
    \midrule
    SFT &There have been thirty-seven managers since 1896. \\
    \midrule
    KD & There have been 28 managers of Arsenalf FC.\\
    \midrule
    SeqKD &Arsenal has had 20 permanent managers, 8 caretaker managers and 2 manager with dead in the job. \\
    \midrule
    ImitKD &There have been thirty-seven managers since 1896. \\
    \midrule
    MiniLLM &Since 1897, Arsenal have had twenty-eight managers. Eight of these were caretaker managers. The most successful manager is Ars\'ene Wener, who won three Premier Leagues, seven FA Cups, and seven Community Shield trophies. He is the longest-serving manager, as he surpassed George Alison's record of twelve years in October 2,009. Two managers have died in this job; Herbert Chapman and Tom Whitaker. \\
    \midrule
    GKD & There have been thirty managers of Arsenals football club since 1896. This includes twenty permanent managers and eight caretaker or interim managers.\\
    \midrule
    DISTILLM &There have been 20 permanent managers and 8 caretaker managers for a total of 28 managers. \\
    \midrule
    Ours & There have been 20 permanent and 8 caretaker managers at Arsenal.\\
    \bottomrule[1.2pt] 
  \end{tabularx}
    \label{case2}
\end{table}

\begin{table}[!ht]
  \caption{Instruction-following cases from the databricks-dolly-15k dataset.}
  \centering
  \begin{tabularx}{\textwidth}{@{}l|X@{}}
    \toprule[1.2pt]
    \multicolumn{2}{c}{Case \#3, Category: brainstorming} \\
    \midrule
    Instruction & What is a good way to get around without a car? \\
    \midrule
    Input & - \\
    \midrule
    Groundtruth & If you live in a city you can access public transport such as buses and trains. You can also walk, use a bicycle, or electric scooter to get around. \\
    \midrule
    \midrule
    SFT &When you don't have a car, you need to find alternative ways to get around. If you are able-bodied and live in a city, your best bet would be to use public transportation. Every city has trains, buses and other modes of transportation that are accessible without a car.If you are unable-bodied or have a car that doesn't work, you may want to consider ride-sharing services such as Uber and Lyft. \\
    \midrule
    KD & A few examples of ways to get around without a personal car are using public transportation such as the bus or subway, carpooling with friends or coworkers, or using ride-sharing services such as Uber or Lyft. If none of these options are viable for you, it is still possible to get around without a vehicle; for example, it is often possible to walk or ride a bike for short distances, and in some cases, it is possible to use a bicycle, scooter, or skateboard for longer distances.\\
    \midrule
    SeqKD &There are multiple options for getting around without a car. Biking, walking, riding a scooter, riding a bike with a scooter, using public transportation, rideshare, or a moped are all good options. If you are in a smaller town where public transportation is available, walking or riding a bike may also be a good option. \\
    \midrule
    ImitKD & Public transportation is a good way to avoid a need for a car. In many cities in the US, public transportation includes buses, subways, and trains that can take you where you need to go. For shorter distances, many cities and towns have bike-shares which can provide an easy way to get around without having to rely on a car. \\
    \midrule
    MiniLLM & There are many ways to get around without a driver's license or car. The most common way is to walk, but you can also use a bicycle, public transportation, or ride-sharing services. If you live in a city, you can also take advantage of public transportation, which is often more affordable than owning a car. \\
    \midrule
    GKD & There are multiple ways to get around without a personal vehicle. Public transportation is available in many cities and towns. Many cities also have bike share programs, where you can rent a bike or low-cost electric scooter for short trips. You can also walk, but this may depend on the distance and weather. \\
    \midrule
    DISTILLM & There are many ways to get around without cars. You can walk, ride a bike, use public transportation, or ride share services such as Uber and Lyft. \\
    \midrule
    Ours & If you don't have a car, there are still many ways to get around. You can take public transportation, which includes buses, trains, and subways. If you are near a large city, you may also have access to light rail, which is like a bus that runs on rails. If you are in a suburb, you can use public transportation such as a bus or train. If you live in a small town, you may need to take a taxi or ride share service. \\
    \bottomrule[1.2pt] 
  \end{tabularx}
    \label{case3}
\end{table}

\clearpage
\newpage

\subsection{Example of the sequence correction and re-generation (SCRG) strategies}
\label{example_SCRG}
\begin{table}[!ht]
  \caption{Example of the student-generated output sequence using the sequence correction and re-generation strategy. The \textcolor{red}{red} token represents the detected position of the error token. }
\begin{center}
	\begin{tcolorbox}[colback=gray!10,
		colframe=black,
		width=14cm,
		arc=1mm, auto outer arc,
		boxrule=0.8pt,
		]
  		\textbf{Instruction:} What is the difference between men's and women's lacrosse
		\newline\newline
		\textbf{Samples from student:} Men's lacrosse has a limited \textcolor{red}{amount} of time to play play play as as as as as as as as as as as as as as as as as as as as
		\newline\newline
		\textbf{Student's Prediction:} Men's lacrosse is a \textcolor{red}{of} of of to play and play play as as as as as as as as as as as as as as as as as as as as
		\newline\newline
		\textbf{Teacher's Prediction:} Men's lacrosse is a \textcolor{red}{limited} number of movesouts play each each. opposed opposed they a they opposed opposed opposed opposed opposed opposed opposed opposed opposed opposed opposed opposed they a they opposed opposed opposed opposed opposed opposed opposed opposed opposed opposed 
            \newline\newline
            \textbf{Re-sample:} Men's lacrosse has a \textcolor{red}{limited} number of players and women's lacrosse has a maximum number of players.
	\end{tcolorbox}
\end{center}
\end{table}

\subsection{Prompt template for the  instruction-following task}

\begin{table}[!ht]
  \caption{The prompt template for training and evaluation of instruction-following task experiments.}
\begin{center}
	\begin{tcolorbox}[colback=gray!10,
		colframe=black,
		width=14cm,
		arc=1mm, auto outer arc,
		boxrule=0.8pt,
		]
    Below is an instruction that describes a task.\\
    Write a response that appropriately completes the request.\\
  
  \#\#\# Instruction:\newline
  \{instruction\}
  \newline

  \#\#\# Input:\newline
  \{input\}
  \newline
  
  \#\#\# Response:\\
\end{tcolorbox}
\end{center}
\label{template}
\end{table}

\end{document}